\documentclass{article}
\usepackage{spconf,amsmath,epsfig,tabularx,amssymb}

\copyrightnotice{978-1-7281-1485-9/20/$31.00 ©2020 IEEE}

\let\OLDthebibliography\thebibliography
\renewcommand\thebibliography[1]{
  \OLDthebibliography{#1}
  \setlength{\parskip}{0pt}
  \setlength{\itemsep}{0pt plus 0.3ex}
}

\newcommand*\rot{\rotatebox{90}}

\begin{document}\sloppy

\title{\vspace{-0.1em}Deep learning classification with noisy labels\vspace{-0.1em}}

\name{Guillaume SANCHEZ, Vincente GUIS, Ricard MARXER, Frédéric BOUCHARA\vspace{-0.2em}}
\address{\texttt{gsanchez@hexaglobe.com;} \\
\texttt{guillaume-sanchez@etud.univ-tln.fr;} \texttt{$\{$guis,marxer,bouchara$\}$@univ-tln.fr}\\
Université de Toulon, Aix Marseille Univ, CNRS, LIS, Marseille, France}

\maketitle

\begin{abstract}
Deep Learning systems have shown tremendous accuracy in image classification, at the cost of big image datasets. Collecting such amounts of data can lead to labelling errors in the training set. Indexing multimedia content for retrieval, classification or recommendation can involve tagging or classification based on multiple criteria. In our case, we train face recognition systems for actors identification with a closed set of identities while being exposed to a significant number of perturbators (actors unknown to our database). Face classifiers are known to be sensitive to label noise. We review recent works on how to manage noisy annotations when training deep learning classifiers, independently from our interest in face recognition.
\end{abstract}
\begin{keywords}
image classifier, noisy dataset, label noise, noisy training
\end{keywords}
\section{Introduction}
\label{sec:intro}

Learning a deep classifier requires building a dataset. Datasets in media are often situation dependant, with different looking sets or landscape or exhibiting various morphologies, even non-human for face recognition, especially in fantasy and sci-fi contexts. It becomes tempting to use search engines to build a dataset or sort large image sets based on metadata and heuristics. Those methods are not perfect and label noise is introduced.


It is widely accepted that label noise has a negative impact on the accuracy of a trained classifier. Several works have started to pave the way towards noise-robust training. The proposed approaches range from detecting and eliminating noisy samples, to correcting labels or using noise-robust loss functions. Self-supervised, unsupervised and semi-supervised learning are also particularly relevant to this task since those techniques require few or no labels.

In this paper we propose a review of recent research on training classifiers on datasets with noisy labels. We will reduce our scope to the data-dependant approaches estimating or correcting the noise in the data. It is worth mentioning that some works aim to make learning robust by designing new loss functions \cite{Bitempered,GeneralizedCE} without inspecting or correcting the noisy dataset in any way. Those approaches are beyond the scope of our study.

We first define label noise and summarize the different experimental setups used in the literature. We conclude by presenting recent techniques that rely on datasets with noisy labels. This work is inspired by \cite{NoiseSurvey}, extending it to deep classifiers.

\section{Overview of techniques}
All the techniques presented will vary in different ways defined and presented briefly in this section. They can differ on the noise model they build upon, and whether they handle open or closed noise, presented in subsection \ref{defs}, and based on \cite{NoiseSurvey}. Those noise models might need some additional human annotations in the dataset in order to be estimated, introduced in subsection \ref{annotations}. Subsection \ref{detection} will shortly enumerate approaches used for noisy samples detection, when needed. Once noisy samples have been detected, they can be mitigated differently, as outlined in subsection \ref{strategies}.

The various combinations taken by the approaches reviewed here are summed up in Table 1.

\subsection{Problem definition}

\label{defs}
\subsubsection{Models of label noise}

In the datasets studied here, we posit that each sample $x_i$ of a dataset has two labels: the true and unobservable label $y_i$, and the actual label observed in the dataset $\hat{y}_i$. We consider the label noisy whenever the observed label is different from the true label. We aim to learn a classifier $f(x_i)$ that outputs the true labels $y_i$  from the noisy labels $\hat{y}_i$. We denote a dataset $D$ as $D = \{(x_0, \hat{y}_0, y_0), ..., (x_n, \hat{y}_n, y_n)\}$. As presented in \cite{NoiseSurvey} the dataset label noise can be modeled in three way in descending order of generality.

1) The most general model is called \textbf{Noise Not At Random} (NNAR). It integrates the fact that corruption can depend on the actual sample content and actual label. It requires complex models to predict the corruption that can be expected.

\begin{eqnarray}
P(\hat{y}=c|x)=\sum_{c' \in C} P(\hat{y}=c|y=c',x)P(y=c'|x)
\end{eqnarray}

2) \textbf{Noise At Random} (NAR) assumes that label noise is independent from the sample content and occurs randomly for a given label. Label noise can be modeled by a confusion matrix that maps each true label to labels observation probabilities. It implies that some classes may be more likely to be corrupted. It also allows for the distribution of resulting noisy labels not to be uniform, for instance in naturally ambiguous classes. In other words, some pairs of labels may be more likely to be switched than others.

\begin{eqnarray}
P(\hat{y}=c|x) &=& P(\hat{y}=c)  \nonumber \\
~ &=& \sum_{c' \in C} P(\hat{y}=c|y=c')P(y=c')
\end{eqnarray}

3) The least general model, called \textbf{Noise Completely at Random} (NCAR), assumes that each erroneous label is equally likely and that the probability of an error is the same among all classes. For an error probability $E$, it corresponds to a confusion matrix with $P(E=0)$ on the diagonal and $P(E=1) / (|C| - 1)$ elsewhere. The probability of observing a label $\hat{y}$ of class $c$ among the set of all classes $C$ is

\begin{eqnarray}
P(\hat{y}=c|x) &=& P(\hat{y}=c)  \nonumber \\
~ &=& P(E=0)P(y=c) \nonumber \\
~ && + P(E=1)P(y \neq c)
\end{eqnarray}

\subsubsection{Closed-set, open-set label noise}

We distinguish \textbf{open-set} and \textbf{closed-set} noise. In closed-set noise, all the samples have a true label belonging to the classification taxonomy. For instance, a chair image is labeled "table" in a furniture dataset. In open-set noise this might not be the case, in the way a chair image labeled "chihuahua" in a dog races dataset has no correct label.

\subsection{Types of additional human annotations}
\label{annotations}

While training is done on a dataset with noisy labels, a cleaned test set is needed for evaluating the performance of the model. Those clean labels can be acquired from a more trusted yet limited source of data or via human correction.

We may also assume that a subset of the training set can be cleaned. A trivial approach in such cases, is to discard the noisy labels and perform semi-supervised learning using the validated ones and the rest of data as unlabeled. In noisy label training, one aims to exploit the noisy labels as well.

We can imagine a virtual metric, the complexity of annotation of a dataset, determined by factors such as the number of classes, the ambiguity between classes and the domain knowledge needed for labelling. A medical dataset could be hard to label even if it has only two classes while a more general purpose dataset could have a hundred classes that can easily be discriminated if they are all different enough. When the dataset is simple, true label correction can be provided without prohibitive costs.  When it is not, a reviewer can sometimes provide a boolean annotation saying that the label is correct or not, which might be easier than recovering the true labels.

A dataset can then provide (1) no annotations, (2) corrected labels or (3) verified labels for a subset of its labels.

\subsection{Detecting the noisy labels}
\label{detection}

When working on a per-sample decision basis, we often perform noisy samples detection. There are several sources of information to estimate the relevance of a sample to its observed label. In the analyzed papers, four families of methods can be identified. Most of them manipulate the classifier learned, either through its performance or data representation.

1) Deep features are extracted from the classifier during training. They are analyzed with Local Outlier Factor (LOF) \cite{LOF} or a probabilistic variant (pLOF). Clean samples are supposed to be in majority and similar so that they are densely clustered. Outliers in feature space are supposed to be noisy samples.

2) The samples with a high training loss or low classification confidence are assumed to be noisy. It is assumed that the classifier does not overfit the training data and that noise is not learned.

3) Another neural network is learned to detect samples with noisy labels.

4) Deep features are extracted for each sample from the classifier. Some prototypes, representing each class, are learnt or extracted. The samples with features too dissimilar to the prototypes are considered noisy.

\subsection{Strategies with noisy labels}
\label{strategies}

Techniques mitigating noise can be divided in 4 categories. One is based on the Noise At Random model, using statistical methods depending only on the observed labels. The three other methods use Noise Not At Random and need a per sample noise evaluation.

1) One can \textbf{re-weight the predictions} of the model with a confusion matrix to reflect the uncertainty of each observed label. This is inherently a closed-set technique as the probability mass of the confusion matrix has to be divided among all labels.

2) Instead of re-weighting the predictions, we can \textbf{re-weight their importance} in the learning process based on the likelihood of a sample being noisy. Attributing a zero weight to noisy samples is a way to deal with open-set noise.

3) Supposedly erroneous samples can be \textbf{unlabeled}. The sample is kept and used differently, through semi-supervised or unsupervised techniques.

4) Finally, we can try to \textbf{fix the label} of erroneous samples and train in a classical supervised way.

\section{Experimental Setups}

While \textbf{CIFAR-10} \cite{CIFAR10} remains one of the most used datasets in image classification due to its small image sizes, relatively small dataset size, and not-too-easy taxonomy, it has clean labels that are unsuitable for our works. CIFAR-10 contains 60000 images evenly distributed among 10 classes such as "bird", "truck", "plane" or "automobile".. It is still largely employed in noisy label training with artificial random label flipping, in a controlled manner to serve whichever method is shown. However, synthetically corrupting labels fails to exhibit the natural difficulties of noisy labels due to ambiguous, undecidable, or out of domain samples. MNIST \cite{MNIST} can be employed under the same protocols, with a reduced size of classes of handwritten digits, each composed of 1000 images.

\textbf{Clothing1M} \cite{MassiveNoisy} contains 14 classes of clothes for 1 million images. The images, fetched from the web, contain approximately 40\% of erroneous labels. The training set contains 50k images with 25k manually corrected labels, the validation set has 14k images and the test set contains 10k samples. This scenario fits our low annotation complexity situation where labels can be corrected without too much difficulty, but the size of the dataset makes a full verification prohibitive.

\textbf{Food101-N} \cite{CleanNet} has 101 classes of food pictures for 310k images fetched from the internet. About 80\% of the labels are correct and 55k labels have a human provided verification tag in the training set. This dataset rather describes the high annotation complexity scenario where the labels are too numerous and semantically close for an untrained human annotator to correct them. However, verifying a subset of them is feasible.

Finally, \textbf{WebVision} \cite{WebVision} was scraped from Google and Flickr in a big dataset mimicking ILSVRC-2012 \cite{ImageNet}, but twice as big. It contains the same categories, and images were downloaded from text search. Web metadata such as caption, tags and description were kept but the training set is left completely uncurated. A cleaned test set of 50k images is provided. WebVision-v2 extends to 5k classes and 16M training images.

When working on image data, all the papers used classical modern architectures such ResNet \cite{Resnet}, inception \cite{inception} or VGG \cite{VGG}.

\begin{table*}[t]
\label{TheTableau}
\begin{tabularx}{\textwidth}{|p{3cm}||c|c|p{1cm}|p{1.5cm}||c|c|c||X|X|X|X||X|X|X|X|}
\hline
~ & \multicolumn{4}{c||}{\textbf{Strategy}} & \multicolumn{3}{c||}{\textbf{Annotations}} & \multicolumn{4}{c||}{\textbf{Detection}} & \multicolumn{4}{c|}{\textbf{Datasets}}\\
\hline
    ~ & \rot{\textbf{Reweight predictions}}\rot{(NAR, Closed-set)} & \rot{\textbf{Reweight or remove samples}}\rot{(NNAR, Open-set)} & \rot{\textbf{Unlabel samples}}\rot{(NNAR, Open-set)} & \rot{\textbf{Fix labels}}\rot{(NNAR, Closed-set)} & \rot{\textbf{No correction}} & \rot{\textbf{Corrected labels}} & \rot{\textbf{Verified labels}} &
    \rot{\textbf{Local Outlier Factor}} & \rot{\textbf{High loss / Low confidence}} & \rot{\textbf{Model}} & \rot{\textbf{Similarity to prototypes}} & \rot{\textbf{CIFAR-10 / MNIST}}\rot{(Synthetic noise)} & \rot{\textbf{Food-101N}}\rot{(Verified labels)} & \rot{\textbf{Clothing1M}}\rot{(Corrected labels)}
    & \rot{\textbf{WebVision}}\rot{(Raw labels)}\\
\hline
\hline

NLNL \cite{NLNL} & ~ & \checkmark & ~ & negative labels & \checkmark & ~ & ~ & ~ & \checkmark & ~ & ~ & \checkmark & ~ & ~ & \\
\hline
Iterative Noise Filtering \cite{IterativeNoiseFiltering} & ~ & ~ &  without \newline entropy loss & with entropy loss & \checkmark & ~ & ~ & ~ & \checkmark & ~ & ~ & \checkmark & ~ & ~ &\\
\hline

(Ren et al, 2018) \cite{Reweight} & ~ & \checkmark & ~ & ~ & \checkmark & ~ & ~ & ~ & ~ & \checkmark & ~ & \checkmark & ~ & ~ & \\
\hline
Iterative learning \cite{IterativeLearning} & ~ & \checkmark & ~ & ~ & \checkmark & ~ & ~ & \checkmark & ~ & ~ & ~ & \checkmark & ~ & ~ & \\
\hline

NLNN \cite{NLNN} & \checkmark & ~ & ~ & \checkmark & \checkmark  & ~ & ~ & ~ & ~ & ~ & ~ & \checkmark & \multicolumn{3}{c|}{\& TIMIT} \\
\hline
(Hendrycks et al, 2018) \cite{Trusted} & \checkmark & ~ & ~ & ~ & ~ & \checkmark & ~ & ~ & ~ & ~ & ~ & \checkmark & \multicolumn{3}{c|}{\& NLP} \\
\hline

Deep Self-Learning \cite{SelfLearning} & ~ & ~ & ~ & \checkmark & \checkmark & ~ & ~ & ~ &  ~ & ~ & \checkmark & ~ & \checkmark & \checkmark &\\
\hline
CleanNet \cite{CleanNet} & ~ & \checkmark & ~ & ~ & ~ & ~ & \checkmark ~ & ~ & ~ & \checkmark & \checkmark & ~ & \checkmark & \checkmark & \\
\hline

(Xiao et al, 2015) \cite{MassiveNoisy} & \checkmark & ~ & ~ & ~ & ~ & \checkmark & ~ & ~ & ~ & \checkmark & ~ & ~ & ~ & \checkmark & \\
\hline
CurriculumNet \cite{CurriculumNet} & ~ & \checkmark & ~ & ~ & \checkmark & ~ & ~ & ~ & \checkmark & ~ &  ~ &  ~ &  ~ &  ~ &  \checkmark \\
\hline

Co-Mining \cite{CoMining} & ~ & \checkmark  & ~ & ~ & \checkmark  & ~ & ~ & ~ & \checkmark & ~ & ~ & \multicolumn{4}{c|}{face rec}\\
\hline
\end{tabularx}
\caption{Approaches according to annotations in the dataset. Notes: TIMIT is a speech to text dataset, "NLP" is a set of natural language processing datasets (Twitter, IMDB and Stanford Sentiment Treebank), "face rec" denotes classical face recognition datasets (LFW, CALFW, AgeDB, CFP)}
\end{table*}

\section{Approaches}
\label{Approaches}

\subsection{Prediction re-weighting}

Given a softmax classifier $f(x_i)$ for a sample $x_i$, prediction re-weighting mostly implies estimating the confusion matrix $C$ in order to learn $C^Tf(x_i)$ in a supervised fashion with the noisy labels. Doing so will propagate the labels' confusion in the supervising signal to integrate the uncertainty about label errors. The main difference between the approaches lies in the way $C$ is estimated.

In \textbf{Noisy Label Neural Networks} \cite{NLNN}, noisy labels are assumed to come from a real distribution observed through a noisy channel. The algorithm performs an iterative Expectation Maximization algorithm. In the Expectation step, correct labels $c_i$ are guessed through $C^Tf(x_i)$ while in the Maximization step, $C$ is estimated from the confusion matrix between guessed labels $c_i$ and dataset labels $\hat{y}_i$. Finally, $f(x_i)$ is trained on guessed labels $c_i$. The process is repeated until convergence.

Taking a more direct approach, \textbf{(Xiao et al, 2015)} \cite{MassiveNoisy} directly approaches $C$ by manually correcting the labels of a subset of the training set. Then, a secondary neural network $g(x_i)$ is defined, giving to each sample a probability $P(z_i | x_i)$ of being (1) noise free, that is $\hat{y}_i = y_i$, (2) victim of completely random noise (NCAR), ie $P(\hat{y}_i | y_i) = (U-I)y_i$ such that the matrix $U$ is uniform and all rows of $U-I$ sums to 1, or (3) confusing label noise (NAR), $P(\hat{y}_i | y_i) = C^T\hat{y}_i$. Finally, $f(x_i)$ is trained on the noisy labels so as to minimize $L_\text{CE}(z_{1i} f(x_i) + z_{2i}(U-I)f(x_i) + z_{3i}C^Tf(x_i), \hat{y}_i)$ with $L_\text{CE}$ the cross entropy loss function.

\textbf{(Hendrycks et al, 2018)} \cite{Trusted} first train a model on the dataset with noisy labels. This model is then tested on a corrected subset and its predictions errors are used to build the confusion matrix $C$. Finally $f(x_i)$ is trained on the corrected subset and $C^Tf(x_i)$ is trained on the noisy subset.

\subsection{Sample importance re-weighting}

For a softmax classifier $f(x_i)$ trained with a loss function such as cross-entropy $L_{\text{CE}}$, sample importance re-weighting consists in finding a sample weight $\alpha_i$ and minimizing $\alpha_iL_\text{CE}(f(x_i), \hat{y}_i)$. For a value $\alpha_i$ close to 0, the example has almost no impact on training. $\alpha_i$ values larger than 1 emphasize examples. If $\alpha_i$ is exactly 0, then it is analogous to removing the sample from the dataset.

\textbf{Co-mining} \cite{CoMining} investigates face recognition where correcting labels is unapproachable for a large number of identities, and most likely a situation of open-set noise. Two neural nets $f_1$ and $f_2$ are given the same batch. For each net, the losses $l_{1i} = L(f_1(x_i), \hat{y}_i)$ and $l_{2i} = L(f_2(x_i), \hat{y}_i)$ are computed for each sample and sorted. The samples with the highest loss for both nets are considered noisy and are ignored. The samples $s_{1i}$ and $s_{2i}$ that have been kept by $f_1$ and $f_2$ are considered clean and informative: both nets agreed. Finally, the samples kept by only one net are considered valuable to the other. Backpropagation is then applied, with clean faces weighted to have more impact, valuable faces swapped in order to learn $f_1$ with $s_{2i}$ and $f_2$ with $s_{1i}$, and low quality samples are discarded.

\textbf{CurriculumNet} \cite{CurriculumNet}
trains a model on the whole dataset. The deep features of each sample are extracted, and from the Euclidean distance between features vectors, a matrix is built. Densities are estimated, 3 clusters per class are found with k-means, and ordered from the most to least populated. Those three clusters are used for training a classifier with a curriculum, starting from the first with weight 1, then the second and third, both weighted $0.5$.

\textbf{Iterative learning} \cite{IterativeLearning} chooses to operate iteratively rather than in two phases like CurriculumNet. The deep representations are analyzed throughout the training with a probabilistic variant of Local Outlier Factor \cite{LOF} for estimating the densities. Local outliers are deemed noisy. The unclean samples importance is reduced according to their probability of being noisy. A contrastive loss working on pairs of images is added to the cross entropy. It minimizes the euclidean distance between the representation of samples considered correct and of the same class, and maximizes the Euclidean distance between clean samples of different class or clean and unclean samples. The whole process is repeated until model convergence.

We can also employ meta-learning by framing the choice of the $\alpha_i$ as values that will yield a model better at classifying unseen examples after a gradient step. \textbf{(Ren et al, 2018)} \cite{Reweight} performs a meta gradient step on $L=\alpha_iL_\text{CE}(f(x_i), \hat{y}_i)$ then evaluate the new model on a clean set. The clean loss is backpropagated back through $L$, for which the gradient $\eta$ gives the contribution of each sample to the performance of the model on the clean set after the meta step. By setting $\alpha_i = \text{max}(0, \eta_i)$, the samples that impacted the model negatively are discarded, and the positive samples get an importance proportional to the improvement they bring.

\textbf{CleanNet} \cite{CleanNet} learns what it means for a sample to come from a given class distribution, utilizing a correct / incorrect tag provided by human annotators. A pretrained model extracts deep features of the whole dataset. Then, they run a per-class K-Means, and find the images with features closest to the centroids as a set $v_c$ ofreference images for that class $c$. A deep model $g(v_c)$ encodes the set into a single prototype. A third deep model $h(x_i)$ encodes the query image $x_i$ in a prototype. We learn to maximize $w_{ci}=\cos(g(v_{c}), h(x_{i}))$ if $x_i$ has a correct label $c$, and to minimize it otherwise. This relevance score is used to weigh the importance of that sample when training a classifier with $\max(0, w_{\hat{y}_i}) L_{\text{CE}}(f(x_i), \hat{y}_i)$.

Instead of getting a consistent wrong information from an erroneous label, \textbf{NLNL} \cite{NLNL} (not to be confused with NLNN) instead samples a label $\Tilde{y}_i \neq \hat{y}_i$ and uses negative learning, a negative cross-entropy version that minimizes the probability of $\Tilde{y}_i$ for $x_i$. As the number of classes grows, the more likely the sampled label $\Tilde{y}_i$ is indeed different of $y_i$ and noise is mitigated, despite being less informative. Then only samples with a label confidence above $1/|C|$ are kept and used negatively in a second phase called Selective Negative Learning (SelNL). Finally, examples with confidence over a high threshold (0.5 in the paper) are used for positive learning with a classical cross entropy and their label $\hat{y}_i$.

\subsection{Unlabeling}

\textbf{Iterative Noise Filtering} \cite{IterativeNoiseFiltering}:
A model is trained on the noisy dataset. An exponential moving average estimate of this model is then used to analyze the dataset. Samples classified correctly are considered clean, while the label is removed. The model is further trained with both a supervised and unsupervised objective for labeled and unlabeled samples. The samples with labels are used with a cross entropy loss. For each unlabeled sample, we maximize $\max_c f(x_i)_c$ in order to reinforce the model's prediction, while maximizing the entropy of the predictions over the whole batch to avoid degenerate solutions. Datasets labels are evaluated again according to the average model. Training restarts with removed and restored labeled. This procedure is repeated while testing convergence improves.

\subsection{Label fixing}

A few methods already listed above try to fix the labels as part of their approach. While listed as a sample re-weighting method, \textbf{NLNL} \cite{NLNL} also employs a sort of label fixing procedure by using the negative labels. Similarly, \textbf{(Bekker and Gold-berger, 2016)} \cite{NLNN} attempts to fix the labels while estimating the confusion matrix. Finally, \textbf{Iterative Noise Filtering} \cite{IterativeNoiseFiltering}, assumes that the class with the highest prediction for the unlabeled examples is correct.

\textbf{Deep Self-Learning} \cite{SelfLearning} learns an initial net on noisy labels. Then, deep features are extracted for a subset of the dataset. A density estimation is made for each class and the most representative prototypes are chosen for each cluster. The similarity of all samples to each set of prototypes is computed to re-estimate correct labels $\Tilde{y}_i$. The model training continues with a double loss balancing learning from the original label or the corrected one $L=\lambda L_\text{CE}(f(x_i), \hat{y}_i) + (1-\lambda)L_\text{CE}(f(x_i), \Tilde{y}_i)$ with hyper-parameter $\lambda \in [0 1]$. We iterate between label correction and weighted training until convergence. Note that contrarily to sample weighting techniques that weigh the contribution of each sample in the loss, all samples have an equal importance, but we place a cursor as a hyper-parameter to balance between contribution from the noisy labels and corrected labels.

\section{Discussion and conclusions}

Those approaches cover a wide variety of use cases, depending on the dataset: whether is has verified or corrected labels or not, and the estimated proportion of noisy labels. They all have different robustness properties: some might perform well in low noise ratio but deteriorate quickly while others might have a slightly lower optimal accuracy but do not deteriorate as much with high noise ratio.

Re-weighting predictions performs better on flipped labels rather than uniform noise as shown in the experiments on CIFAR-10 in \cite{Trusted}. As noise becomes close to a uniform noise, the entropy of the confusion matrix $C$ increases, labels provide more diffused information, and prediction re-weighting is less informative. CIFAR-10 being limited to 10 classes, NLNN \cite{NLNN} is shown to scale with a greater number of classes on TIMIT. However those approaches only handle closed-set noise by design, and while adding an additional artificial class for out-of-distribution samples can be imagined, none of the works reviewed here explored this strategy.

Noisy samples re-weighting scales well: \cite{CurriculumNet} scales in number of samples and classes as the experiments on WebVision shows, \cite{CoMining} is able to scale to face recognition datasets and open-set noise at the expense of training two models, CleanNet generalizes its noisy samples detection by manually verifying a few classes.

However, NLNL \cite{NLNL} may not scale as the number of classes grows: despite having negative labels that are less likely to be wrong, they also become less informative. 

We can expect unlabeling techniques to grow as the semi-supervised and unsupervised methods gets better, since any of those can be used once a sample had its label removed. One could envision utilizing algorithms such as MixMatch \cite{MixMatch} or Unsupervised Data Augmentation \cite{UDA} on unlabeled samples.

Similarly, the label fixing strategies could benefit from unsupervised representation learning to learn prototypes that makes it easier to discriminate hard samples and incorrect samples. Deep self-learning \cite{SelfLearning} is shown to scale on Clothing1M and Food-101N. It would be expected however that those approaches become less accurate as the number of classes grows or the classes get more ambiguous. Some prior knowledge or assumptions about the classes could be used explicitly by the model. Iterative Noise Filtering \cite{IterativeNoiseFiltering} in its entropy loss assumes that all the classes are balanced in the dataset and in each batch.

Training a deep classifier using a noisy labeled dataset is not a single problem but a family of problems, instantiated by the data itself, noise properties, and provided manual annotations if any. As types of problems and solutions will reveal themselves to the academic and industrial deep learning practitioners, deciding on a single metric, a more thorough and standardized set of tests might be needed. This way, it will be easier to answer questions about the use of domain knowledge, generality, tradeoffs, strengths and weaknesses, of noisy labels training techniques depending on the use-case.

In the face recognition system, that we are building, label noise have varying causes: persons with similar names; confusion with lookalikes; related persons that appear together; erroneous faces detected on signs or posters in the picture; errors from the face detector that are not faces; and random noise. All those situations represent label noise with different characteristics and properties that must be handled with those algorithms. We believe those issues are more general than this scenario and find an echo in the broader multimedia tagging and indexing domain. 

\bibliographystyle{IEEEbib}
\bibliography{icme2020template}

\end{document}